\documentclass[lettersize,journal]{IEEEtran}
\usepackage{amsmath,amsfonts}
\usepackage{algorithmic}
\usepackage{algorithm}
\usepackage{array}
\usepackage[caption=false,font=normalsize,labelfont=sf,textfont=sf]{subfig}
\usepackage{textcomp}
\usepackage{stfloats}
\usepackage{url}
\usepackage{verbatim}
\usepackage{graphicx}
\usepackage{cite}
\usepackage{multirow}
\usepackage{multicol}
\usepackage{color}
\usepackage{yhmath}
\usepackage{bm}
\usepackage{pifont}
\usepackage{balance}
\usepackage{epsfig}
\usepackage[colorlinks,linkcolor=red]{hyperref}

\usepackage{tikz,xcolor,hyperref}
\definecolor{lime}{HTML}{A6CE39}
\DeclareRobustCommand{\orcidicon}{%
    \begin{tikzpicture}
    \draw[lime, fill=lime] (0,0) 
    circle [radius=0.16] 
    node[white] {{\fontfamily{qag}\selectfont \tiny ID}};    \draw[white, fill=white] (-0.0625,0.095) 
    circle [radius=0.007];    \end{tikzpicture}
    \hspace{-2mm}}
\foreach \x in {A, ..., Z}{%
    \expandafter\xdef\csname orcid\x\endcsname{\noexpand\href{https://orcid.org/\csname orcidauthor\x\endcsname}{\noexpand\orcidicon}}
    }

\begin{document}

\title{JNMR: Joint Non-linear Motion Regression for Video Frame Interpolation}

\author{Meiqin Liu\textsuperscript{*}\orcidB{}, Chenming Xu\textsuperscript{*}\orcidE{}, Chao Yao\textsuperscript{\dag}\orcidC{}, Chunyu Lin\orcidD{}, ~\IEEEmembership{Member,~IEEE}, and Yao Zhao\orcidA{}, ~\IEEEmembership{Fellow,~IEEE}
\thanks{
* {Equal Contributions.}

\dag{ }{Corresponding Author: Chao Yao.}

This work was supported in part by the National Key R\&D Program of China 2022ZD0118001, the National Natural Science Foundation of China under Grant 62372036, 61972028, 62332017 and 62120106009. 

Meiqin Liu, Chenming Xu, Chunyu Lin, and Yao Zhao are with the Institute of Information Science, Beijing Jiaotong University, Beijing 100044, China, and also with the Beijing Key Laboratory of Advanced Information Science and Network Technology, Beijing 100044, China (e-mail: \url{mqliu@bjtu.edu.cn}; \url{chenming_xu@bjtu.edu.cn}; \url{cylin@bjtu.edu.cn}; \url{yzhao@bjtu.edu.cn}).

Chao Yao is with the School of Computer \& Communication Engineering, University of Science and Technology Beijing, Beijing 100083, China (e-mail: \url{yaochao@ustb.edu.cn}).
}
}




\markboth{IEEE TRANSACTIONS ON IMAGE PROCESSING}%
{Shell \MakeLowercase{\textit{et al.}}: A Sample Article Using IEEEtran.cls for IEEE Journals}


\maketitle

\begin{abstract}
Video frame interpolation (VFI) aims to generate predictive frames by motion-warping from bidirectional references. Most examples of VFI utilize spatiotemporal semantic information to realize motion estimation and interpolation. However, due to variable acceleration, irregular movement trajectories, and camera movement in real-world cases, they can not be sufficient to deal with non-linear middle frame estimation. In this paper, we present a reformulation of the VFI as a joint non-linear motion regression (JNMR) strategy to model the complicated inter-frame motions. Specifically, the motion trajectory between the target frame and multiple reference frames is regressed by a temporal concatenation of multi-stage quadratic models. Then, a comprehensive joint distribution is constructed to connect all temporal motions. Moreover, to reserve more contextual details for joint regression, the feature learning network is devised to explore clarified feature expressions with dense skip-connection. Later, a coarse-to-fine synthesis enhancement module is utilized to learn visual dynamics at different resolutions with multi-scale textures. The experimental VFI results show the effectiveness and significant improvement of joint motion regression over the state-of-the-art methods. The code is available at \url{https://github.com/ruhig6/JNMR}.
\end{abstract}

\begin{IEEEkeywords}
Video frame interpolation, multi-variable non-linear regression, motion estimation, interpolation modeling, deformable convolution.
\end{IEEEkeywords}

\section{Introduction}
\IEEEPARstart{T}{he} purpose of video frame interpolation (VFI) is to generate new middle frames from existing reference frames. It is essential for various applications, such as slow-motion generation~\cite{Jiang_2018_CVPR}, frame compensation in video compression~\cite{hu2021fvc,lu2019dvc,pourreza2021extending,Choi2020Deep}, frame recovery~\cite{bao2018high,wu2015modeling}, etc. Typically, high-level global motions and subtle variations of the synthesis frame have the same importance. {Therefore, it is challenging to accurately estimate the complicated motions for frame interpolation modeling}.
\begin{figure}
\centering{
\includegraphics[width=\linewidth]{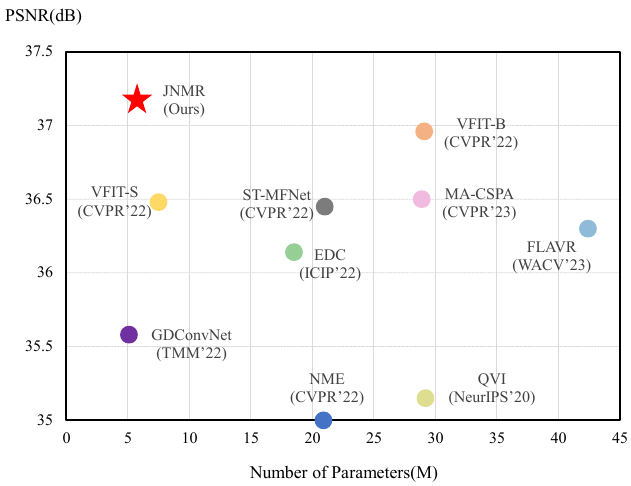}
}
\caption{Quantitative comparison of performance and model parameters with the Vimeo-90K Septuplet dataset. JNMR outperforms state-of-the-art methods with a competitive model size.}
\label{fig1}
\end{figure}

In general, many methods formulate complicated VFI motion estimation as seeking the warping correlation from historical consecutive frames. Optical flow-based solutions~\cite{sun2018pwc,wang2018video} and kernel-based methods~\cite{Xue2019Visual,niklaus2017video} are commonly applied to realize the pipeline. {Optical flow-based methods initially entail utilizing convolutional networks~\cite{dosovitskiy2015flownet,xue2019video} to estimate pixel-level motions. Whereas, the interpolation performance is limited by the accuracy of optical flow estimation.} {Therefore, some works~\cite{Park_2021_ICCV,huang2020rife,bao2019DAIN,park2020robust} propose to utilize exceptional features and supervision operations for improving the flow estimation accuracy.} {However, these methods suffer from high computational costs. Additionally, it is inevitable that the estimation precision of long-term variations and complex occlusions will be decreased by only exploring spatiotemporal dynamics in low-dimensional space. Kernel-based methods~\cite{niklaus2017video,cheng2020video, lee2020adacof,ding2021cdfi} entail utilizing a deformable convolution operation for motion estimation, which can enable the adaption of the spatial dynamic scenes and the expansion of the receptive field of motion. However, the interpolation performance is still limited by the finite relations between inter-frame.}

{To explore temporal correlations, some approaches~\cite{kalluri2023flavr,danier2022enhancing,danier2022spatio,Shi2022VideoFrame,shi2022video} entail extending the motion estimation methods to multiple reference frames interpolation. {These methods primarily address the spatiotemporal correlations of complex motions by focusing on feature synthesis. Certain approaches~\cite{danier2022spatio,Shi2022VideoFrame,shi2022video} gradually divide temporal motions into smaller groups to enhance the accuracy of motion estimation. Other methods~\cite{kalluri2023flavr,danier2022enhancing} utilize 3D convolution to synthesize frames not only in the space dimension but also in the time dimension. However, the above methods mostly involve an assumption that there are uniform motions between consecutive frames according to a linear distribution (as with the black model in Fig.\ref{fig2}(a)) in kinematics.} This assumption fails to consider motion correlations in the temporal dimension of multi-variable regression. To address this problem, some works \cite{xu2019quadratic,liu2020enhanced,zhang2020video, xing2021flow} explore a quadratic interpolation model for multi-variable regression, allowing prediction with variable velocity and non-linear attributes. Furthermore, \emph{Saikat et al.}~\cite{Saikat2022Non} use the coefficients to adaptively select a linear or quadratic model for non-linear motion formulation. \emph{Shen et al.}~\cite{shen2020video} introduce ConvLSTM to combine consecutive linear features as a quadratic regression. {As illustrated in the blue model of Fig.\ref{fig2}(a), these methods entail the prediction of intermediate motions with a curvilinear trajectory instead of the linear geometric center estimation.} 
 
\begin{figure}
\centering{
\includegraphics[width=\linewidth]{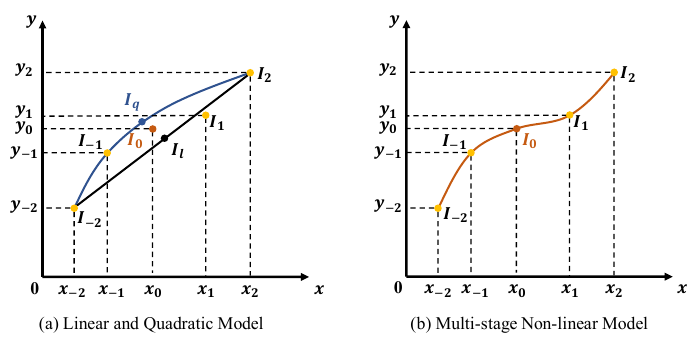}
}
\caption{Visualization of linear/quadratic estimation models and the proposed joint multi-stage non-linear regression model. {(a) $I_l$ denotes the symmetric midpoint referred to as a linear function, and $I_q$ represents the middle motion of the quadratic trajectory. It can be observed that both the linear  (black) and the quadratic (blue) models cannot precisely estimate some relatively complicated intermediate motions in the sequence $\{I_{-2}, I_{-1}, I_{0}, I_{1}, I_{2}\}$. (b) The interpolation is reformulated as decoupling the trajectory into individual regression models. The successive estimation with individual models can be regressed as a reintegration to predict $I_0$.}}
\label{fig2}
\end{figure}

{Nevertheless, the non-linear movements in real-world scenes are always more complicated than previous assumptions due to variable acceleration, irregular movement trajectories, camera movement and shaking, etc. The linear and quadratic models cannot precisely describe non-linear situations. Therefore, in contrast to other methods, we utilize a multi-stage non-linear model to optimize the motion compensation. As shown in Fig.\ref{fig2}(b), VFI is reformulated as the joint decoupled quadratic trajectories to refine the simulation of irregular movements.}

Specifically, we propose a joint non-linear motion regression (JNMR) strategy to optimize the interpolation modeling, where both spatial deformations and temporal variations are considered. The movements can be regarded as a temporal concatenation of multi-stage quadratic models to address complicated motion reconstruction. {ConvLSTM is implemented to initialize existing motions as a distribution in the temporal dimension. To preserve the moderate semantics of variations, we explore a hierarchical network structure before regression to retain the high-level variations in feature learning. Meanwhile, the skip-connection operation between the convolutional layers is utilized to compensate for the subtle variations in the final features}. A coarse-to-fine synthesis enhancement module is designed for different spatial resolution feature synthesis with joint regression to further remove artifacts and unreliable pixels. Experimental results demonstrate the effectiveness of JNMR on video frame interpolation, outperforming state-of-the-art methods, as shown in Fig.\ref{fig1}. In summary, our contributions are as follows:
\begin{itemize}
\item {{We propose a joint multi-variable non-linear motion regression strategy for motion compensation of VFI. The refined interpolation modeling can reconstruct the actual kinematic characteristics among large and complicated motions.}}
\item {{We leverage ConvLSTM to realize the joint multi-stage quadratic model for JNMR, which enhances the temporal consistency in inter-frame prediction.}}
\item We design the regression-driven feature learning module and coarse-to-fine enhancement module, separately optimizing the feature extraction with competitive parameters and pleasing visual qualities on different resolutions.
\end{itemize}

\section{Related Works}

Intermediate frames are generated by motion estimation and interpolation modeling. {Motion estimation methods mainly incorporate optical flow-based methods and kernel-based methods.} {The non-linear interpolation strategy is proposed to match the curve trajectory.} We briefly review related works in the following section. 

\subsection{Optical Flow-based Methods}

{Optical flow-based methods aim to exploit the pixel-wise corresponding relationships of bidirectional reference frames as motion vectors for VFI.} \emph{Xue et al.}~\cite{xue2019video} first introduce bidirectional optical flows for motion estimation. \emph{Park et al.}~\cite{Park2020BMBC} recursively warp the optical flow by tracking the position with the greatest correlation. \emph{Cheng et al.}~\cite{cheng2019multi} utilize a feature pyramid to achieve multi-scale optical flow estimation. \emph{Huang et al.}~\cite{huang2020rife} design a privileged distillation scheme optimized by the ground truth for precisely training intermediate flow models. {\emph{Kong et al.}~\cite{kong2022ifrnet} adopt a gradually refined intermediate feature to efficiently compensate for contextual details.}  \emph{Lu et al.}~\cite{lu2022vfiformer} leverage Transformer to extend the receptive field of optical flow for long-term dependency optimization. To further improve the accuracy of flow estimation, some methods~\cite{Park_2021_ICCV,niklaus2020softmax,hu2022many} have entailed inferring the additional information during the warping of original frames, such as with asymmetric fields and softmax splatting. {\emph{Sim et al.}~\cite{sim2021xvfi} propose a recursive multi-scale structure for extreme VFI on 4K videos.} In addition, many solutions have utilized multi-frame as input to enhance the logical continuity of optical flow. Specifically, \emph{Shen et al.}~\cite{shen2020video} introduce ConvLSTM to estimate optical flow and combine temporal and spatial data. \emph{Park et al.}~\cite{park2020robust} promote exceptional motion detection to improve the robustness of interpolation. These works have achieved state-of-the-art performance in realistic intermediate frame generation. However, the methods involved suffer from inaccurate pixel shifts, ghosting or blurry artifacts that are inevitably generated where the temporal dynamics dependencies are significant for the input frames.

\subsection{Kernel-based Methods}
{Instead of relying on optical flow, kernel-wise convolutional operations have been introduced to expand the receptive field of motion estimation with a lower computation burden.} \emph{Niklaus et al.}~\cite{niklaus2017video} are the first to provide useful insights on how to match the visual scenes and complicated motion for new frame interpolation through a dynamic network. To adapt to inter-frame motion properties, \emph{Lee et al.}~\cite{lee2020adacof} design the AdaCoF model to complement the offset vector coefficient for kernel estimation. \emph{Cheng et al.}~\cite{cheng2021multiple} propose an enhanced deformable separable convolution to estimate relatively complete kernel motions with a non-local neighborhood. To further extend the receptive field of kernel estimation, \emph{Shi et al.}~\cite{Shi2022VideoFrame} and \emph{Tian et al.}~\cite{Tian2022VideoFI} explore a suitable distribution of reference points to effectively train the generalized deformable convolution for capturing the shape of the object. {With regard to the improvement on the concrete network structure of the feature extraction}, \emph{Ding et al.}~\cite{ding2021cdfi} deploy the feature pyramid and network compression in motion learning to strengthen the robustness with parameter reduction. \emph{Wijma et al.}~\cite{wijma2021multi} implement spatial pyramids to increase the receptive field of the model to adaptively adjust the kernel size. Nevertheless, the limitation on temporal dependency still exists in kernel-level motion estimation. Thus, \emph{Choi et al.}~\cite{choi2020channel} attempt to utilize channel attention to distribute individual feature maps for motion extraction with spatiotemporal information. \emph{Kalluri et al.}~\cite{kalluri2023flavr} and \emph{Danier et al.}~\cite{danier2022enhancing,danier2022spatio} adopt multi-scale 3D convolution to solve the spatiotemporal dependence with multiple frames. To integrate the advantages of the optical flow-based method in detailed motions, \emph{Bao et al.}~\cite{Bao2021MEMC} propose the MEMC-Net to estimate motions by optical flow and deformable convolution. \emph{Hu et al.}~\cite{Hu2021Capturing} propose a recurrent motion-enhanced interpolation network based on MEMC-Net to further capture spatiotemporal perspectives. {\emph{Shi et al.}~\cite{shi2022deep} utilize the stacked optical flow estimator, trainable feature extraction and residual convolutional network to increase the quality of the interpolation view.} {\emph{Xiao et al.}~\cite{xiao2022space} propose a joint framework of flow and kernel estimation to achieve the space-time super-resolution of satellite videos.} However, these methods use the compositions of convolutional layers to enable the learning of the spatiotemporal motions in complicated scenes without considering the consecutive variation recurrence. More importantly, both flow-based and kernel-based methods explicitly or implicitly assume linear motion among input frames, which leads to insufficient exploration of higher-order information in motion estimation.

\subsection{Interpolation Modeling Methods}
To further model long-term dynamics consistency from historical observations, some researchers have made innovations in terms of interpolation modeling based on existing motions. {Typically, \emph{Zhou et al.}~\cite{Zhou2022Exploring} develop a texture consistency loss to ensure that interpolated content maintains structures similar to corresponding counterparts.} \emph{Argaw et al.}~\cite{argaw2022long} conduct motions in the same direction as references for current interpolation when there is a large gap between input frames. {In addition, some methods~\cite{chan2021basicvsr,chan2022basicvsr++,zhou2022revisiting} propose the progressive non-linear motion estimation strategy, which aims to approximate complex motions. BasicVSR~\cite{chan2021basicvsr} and BasicVSR++~\cite{chan2022basicvsr++} use temporal alignment for long-term motion formulation in video super-resolution with all frames known, where the accumulation of errors may affect the final performance. \emph{Zhou et al.}~\cite{zhou2022revisiting} propose an iterative alignment strategy that performs gradual refinement for shared sub-alignments. These methods focus on identifying how a finer motion for alignment can be learned with all frames known and on matching motion estimation in temporal dimensionality.} To obtain reliable motions from a kinematic perspective, non-linear models~\cite{xu2019quadratic,liu2020enhanced,zhang2020video,xing2021flow,Saikat2022Non,tulyakov2022time} have been utilized with the aim of approximating complex motion in the real world and overcoming the temporal limitation of camera sensors. \emph{Xu et al.}~\cite{xu2019quadratic} first define a quadratic model against existing linear models. To achieve precise motions, \emph{Liu et al.}~\cite{liu2020enhanced} adopt a rectified quadratic flow prediction formulation with a least squares function. \emph{Zhang et al.}~\cite{zhang2020video} further derive a general curvilinear motion trajectory formula on reference frames without temporal priors. {\emph{Saikat et al.}~\cite{Saikat2022Non} utilize the predicted non-linear flow as coefficients to automatically select linear or quadratic models through space-time convolution networks.} \emph{Tulyakov et al.}~\cite{tulyakov2022time} realize continuous non-linear motion estimation by combining both images and intermediate events. However, due to the incomplete consideration of complicated models in real scenes, the performance of VFI is still limited. {Hence, we reformulate VFI as a joint regression strategy and further expand upon the existing  models to adapt them to kinematic situations.}

\section{Methodology}
\label{sec3}

\subsection{Formulation}
\label{seca}
{Multi-reference VFI entails the assumption that an input dynamic video sequence $\{I_n~|~n\in\{-\frac{N}{2},...,-1,1,...\frac{N}{2}\}\}$ consists of $N$ frames along the time dimension, where each frame is recorded in a spatial region of $\mathbb{R}^{C \times H \times W}$.} The goal is to predict the most likely frame ${I}_0$ for the ground truth intermediate frame $I_{gt}$.

{Following the kernel-based motion estimation methods~\cite{lee2020adacof}, the variations of the input sequence $I_n$ can be modeled as the motions.} Specifically, the input synthesized features can be decoupled into motion vectors using deformable convolution kernels, where the input frames $I_n$ are refined to the warped frames $\hat{I}_n$, as:
\begin{equation}
\hat{I}_n(i,j)=\sum_{p=0}^{K-1}\sum_{q=0}^{K-1}\bm{W}_{p,q}(i,j)I_n(i+dp+\bm{\alpha}_{p,q},j+dq+\bm{\beta}_{p,q})
\label{eq1}
\end{equation}
where $\hat{I}_n(i,j)$ denotes the target pixel at $(i,j)$ in ${I}_n$, $d\in\{0,1,2,...\}$ indicates the dilation of the motion vectors bias $(p,q)$, and $K$ is the kernel size. $\bm{W}_{p,q}$ and $(\bm{\alpha}_{p,q},\bm{\beta}_{p,q})$ denote the kernel weight and motion vectors between $I_n(i,j)$ and $\hat{I}_n(i,j)$.

{Next, considering the occlusion between bidirectional resource frames, the interpolated frame $\hat{I}_{0}$ is formulated as:}
\begin{equation}
\hat{I}_0=O \cdot \sum_{t=-N/2}^{-1}\hat{I}_{n} + (1-O) \cdot \sum_{t=1}^{N/2}\hat{I}_{n}
\label{eq2}
\end{equation}
where $[\cdot]$ is the inner product between two matrices. $O\in[0,1]$ indicates the occlusion generated by the deconvolutional layers with a sigmoid function. However, the spatial information modeled by the above formulation is limited to linear correlations, which neglects the essential information about temporal variable dynamics in real complicated kinematics. 

{To estimate the complicated motion of the intermediate frame, we first reformulate the VFI as a motion-time model:} 
\begin{equation}
\bm{\tilde{y}}_n  ={\bm{\omega}}_n \bm{{x}}_n
\label{eq3}
\end{equation}
{where $\bm{{x}}_n$ denotes the temporal variables. $\bm{\tilde{y}}_n$ is the corresponding predicted motion. $\bm{\omega}_n$ is a regression coefficient related to temporal variables $\bm{M}_n$. Typically, $\bm{M}_{n}$ with the common component $(\bm{W}_n,\bm{\alpha}_n,\bm{\beta}_n)$ is defined as an image wise motion from $I_n \rightarrow I_0$. To further concretize the correlations in Eq.\ref{eq3}, the motions can be commonly understood as the distance between two frames $I_0$ and $I_n$.}

Therefore, according to the general kinematic regularity, motions can be quantified by the instantaneous velocity $\bm{v}_0$ of ${I}_{n}$ and acceleration $\bm{a}_t$ as a distance-time function:
\begin{equation}
\bm{M}_{n+1}-\bm{M}_{n} = \int_0^t(\bm{v}_0+\int_0^k \bm{a}_t dt)dk
\label{eq4}
\end{equation}
{where $(\bm{M}_{n+1}-\bm{M}_{n})$ is a variation vector and represents the distance between two frames. $t$ is the differentiable variable between two adjacent motions. $k$ is the intermediate variable used for integration.} Since $\bm{v}_0$ and $\bm{a}_t$ are difficult to calculate, two adjacent motions cannot determine the kinematic regression model. {Thus, an additional motion $\bm{M}_{n-1}$ is introduced, and the new condition can be formulated as:}
{\begin{equation}
\bm{M}_{n}-\bm{M}_{n-1} = \int_0^t(\bm{v}_0-\int_0^k \bm{a}_t dt)dk
\label{eq0}
\end{equation}}

{Since the instantaneous velocity $\bm{v}_0$ is constant, the kinematic model can be solved as the difference between Eq.\ref{eq4} and Eq.\ref{eq0}:}
\begin{equation}
(\bm{M}_{n+1}-\bm{M}_{n}) - (\bm{M}_{n}-\bm{M}_{n-1}) = 2 \int_0^t \bm{a}_t tdt
\label{eq5}
\end{equation}
{It is evident that the quadratic model can be determined by at least three consecutive motions $\{\bm{M}_{n-1}, \bm{M}_{n}, \bm{M}_{n+1}\} $. {Obviously, this assumption is based on the prior that the movement follows a uniform acceleration.} In real-world scenarios, objects do not always move regularly at a consistent velocity. It is inaccurate to describe the overall motion only by this dynamical model. }

Consequently, we design a general kinematic model to capture complicated motions by combining consecutive independent quadratic models, which can be regarded as uniformly variable motion. The parameters of each individual quadratic model can be defined by the existing motions, and the empirical regression equation of the kinematic model can be determined as:
\begin{equation}
\bm{\hat{y}}_n  =\hat{\bm{\omega}}_n \bm{{x}}_n 
\label{eq6}
\end{equation}
with
\begin{equation}
\begin{split}
    \hat{\bm{y}}_n&=\hat{\bm{M}}_{n}, \\
    \hat{\bm{\omega}}_n&=[\bm{M}_{n},
    \hat{\bm{v}}_{n},\hat{\bm{a}}_{n}], \\
    \bm{{x}}_n&=[1,t,t^2 ]^T
\end{split}
\label{eq7}
\end{equation}
where $\bm{\hat{M}}_{n}$ denotes the individual regressed motions from $\bm{M}_n$. $\hat{\bm{v}}_{n}$ and $\hat{\bm{a}}_{n}$ are the initial velocity and acceleration of ${I}_n$. 

As depicted in Fig.\ref{fig2}(b), the individual quadratic model is inadequate to accurately capture the complicated non-linear motions with irregularity. To alleviate this limitation, we decompose the overall motion into a multi-stage quadratic model. Specifically, three consecutive frames are utilized to form a complete model after regressing the sub-distribution. {Following this pipeline, the whole regression can be defined as the temporal-aware combination of multiple dependent quadratic models as follows:
\begin{equation}
H_{\bm{\theta}} (\bm{\hat{y}}_n)=\bm{\theta} \otimes \bm{Y}
\label{eq8}
\end{equation}
with
\begin{equation}
\begin{split}
    \bm{\theta}&=[\theta_{-\frac{N}{2}+1}, \cdots ,\theta_{\frac{N}{2}-1}], \\
    \bm{Y}&=[\bm{\hat{M}}_{-\frac{N}{2}+1}, \cdots ,\bm{\hat{M}}_{\frac{N}{2}-1} ]^T
\end{split}
\label{eq9}
\end{equation}
{where $H_{\bm{\theta}}(\bm{\hat{y}}_n)$ represents the temporal concatenation of quadratic models, which can be utilized to predict motions. $\bm{\theta}$ denotes the polynomial coefficient collection between different quadratic models. $\otimes$ represents the temporal concatenation instead of linear combination in general matrix multiplication. $\bm{Y}$ contains the individual regressed motions.} It is noted that the above formulation releases the constraint of velocity and acceleration and achieves the connection of the multi-stage quadratic models. {Furthermore, the empirical model could be trained with forward and backward regression:}
\begin{equation}
H_{\bm{\theta}} (\bm{\hat{y}}_n) =\bm{\hat{\theta}} \otimes \bm{\hat{Y}}
\label{eq10}
\end{equation}
{with}
\begin{equation}
\begin{split}
    &\bm{\hat{\theta}}=[\hat{\theta}_{f},\hat{\theta}_{b}], \\
    &\bm{\hat{Y}}=[\hat{\bm{M}}_{f},\hat{\bm{M}}_{b}]^T
\end{split}
\label{eq11}
\end{equation}
{where $\bm{\hat{\theta}}$ which contains $[\hat{\theta}_{f},\hat{\theta}_{b}]$ denotes the collection of bidirectional regressed polynomial coefficients. $\bm{\hat{Y}}$ denotes the second-order regressed motion sequence.} $\hat{\bm{M}}_{f}$ and $\hat{\bm{M}}_{b}$ denote the forward and backward regressed motions, respectively, in a minimal unilateral neighborhood of the intermediate moment. The intermediate instantaneous motion obtained in Eq.\ref{eq11} cannot be directly transformed into the visual location of the final frame. Consequently, the visual movement offset $\Delta\hat{I}_0$ needs to be inferred  from the most adjacent frames $I_{-1}$ and $I_1$, as follows:
\begin{equation}
\Delta\hat{I}_0 = \hat{\theta}_{f} \cdot \varphi(I_{-1}, \hat{\bm{M}}_{f}) + \hat{\theta}_{b} \cdot \varphi(I_{1},\hat{\bm{M}}_{b}) 
\label{eq12}
\end{equation}
where $\hat{\theta}_{f}$ and $\hat{\theta}_{b}$ denote the predicted polynomial coefficients for forward and backward motions. $\varphi$ indicates the warping operation on the reference frame illustrated in Eq.\ref{eq1}. The current predicted frame $\tilde{I}_{0}$ can be incorporated to make up the limitation of long-term dynamics dependency, as:
\begin{equation}
\tilde{I}_{0}=\hat{I}_{0} + \Delta\hat{I}_0
\label{eq13}
\end{equation}
where $\Delta\hat{I}_0$ indicates the visual movement offset, and $\hat{I}_0$ denotes the basic synthesis frame in Eq.\ref{eq2}.

\begin{figure*}[!t]
\centering{
\includegraphics[width=\textwidth]{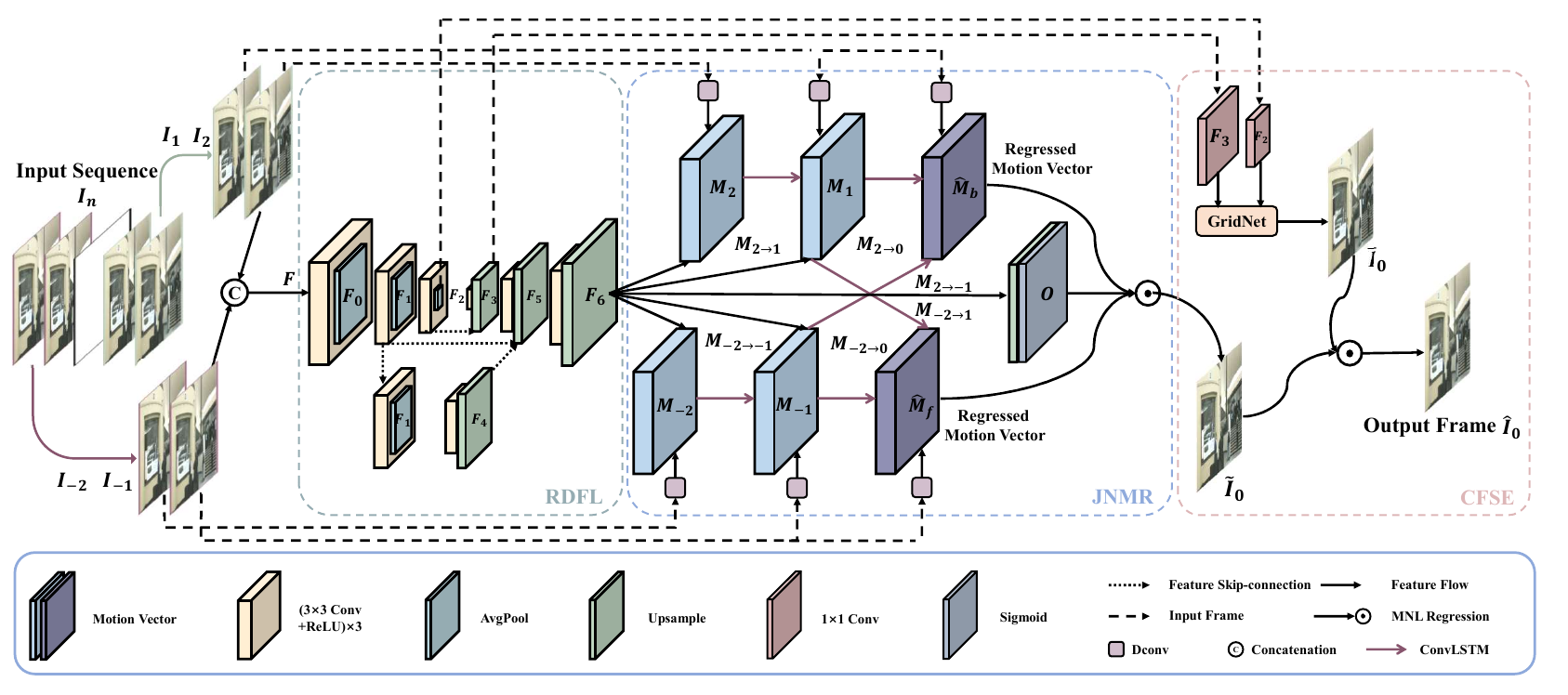}
}
\caption{Illustration of our pipeline for interpolation. In particular, we implement a concatenation of four reference frames $\{I_{-2},I_{-1},I_{1},I_{2}\}$ as the input into the network for regression-driven feature learning. {The fusion feature $\bm{F}_6$ is processed by deformable convolution to obtain the corresponding motion vectors. Then, a joint non-linear regression strategy is proposed to achieve robust interpolation modeling, considering the real kinematic model.} The details are shown in Fig.\ref{fig4}. Finally, coarse-to-fine synthesis enhancement is designed to overcome artifacts caused by complicated motion estimation. }
\label{fig3}
\end{figure*}

\subsection{Regression-Driven Feature Learning}
\label{secc}
Feature learning module can obtain an aggregated feature from the concatenation of input sequences before regression. It is essential for each spatiotemporal dynamic feature to be learned separately in the temporal dimension from the input frame concatenation. Previously developed algorithms, such as CNN-based~\cite{lee2020adacof,ding2021cdfi,kalluri2023flavr} and Transformer-based methods~\cite{liu2020convtransformer,shi2022video}, only formulate feature learning as a general feature filter without consideration given to the internal structure of fusion features. {These stacked filter layers bring structural redundancy and result in the loss of local-global expressions}. Therefore, we propose a regression-driven feature learning (RDFL) network, which is shown in Fig.\ref{fig3}. The RDFL network not only compensates with multiple hierarchical spatial structures to exploit high-level motions with appropriate motion features for regression but also simultaneously reduces the number of parameters.

In detail, the concatenation feature $\bm{F}$ of the multiple candidate frames $\{I_{-2},I_{-1},I_{1},I_{2}\}$ is input into three convolutional layers with the down-sampling operation to obtain the fundamental feature $\bm{F}_0$ for computation reduction. Then, three layers of the hierarchical spatial structure are utilized with $\bm{F}_0$ to extract regression-aware detailed features $\bm{F}_1$ and $\bm{F}_2$ with different resolutions, which can be expressed as:
\begin{equation}
\begin{aligned}
\bm{F}_{0}&=\downarrow(\phi_{conv}(\bm{F})), \\
\bm{F}_{1}&=\downarrow(\phi_{conv}(\bm{F}_0)), \\
\bm{F}_{2}&=\downarrow(\phi_{conv}(\bm{F}_1))
\label{eq19}
\end{aligned}
\end{equation}
where $\downarrow ( )$ denotes the down-sampling operation with average pooling and $\phi_{conv} ( )$ represents three consecutive convolutional layers.

Then, to further enhance the spatial feature expression, a multi-scale fusion strategy is implemented in the hierarchical spatial structures by skip-connection operation. The deconvolutions with the up-sampling operation are adopted to extract features $\bm{F}_3$, $\bm{F}_4$ and $\bm{F}_5$, as follows:
\begin{equation}
\begin{aligned}
\bm{F}_3 &= \uparrow(\phi_{deconv}(\bm{F}_2)) + \bm{F}_1, \\
\bm{F}_4 &= \uparrow(\phi_{deconv}(\bm{F}_1)), \\
\bm{F}_5 &= \uparrow(\phi_{deconv}(\bm{F}_3)) + \bm{F}_0 + \bm{F}_4
\label{eq20}
\end{aligned}
\end{equation}
where $\uparrow ( )$ denotes the up-sampling operation by bilinear interpolation and $\phi_{deconv} ( )$ represents three consecutive deconvolutions. After that, $\bm{F}_6$ is sampled by $\bm{F}_5$, as follows:
\begin{equation}
\bm{F}_6 = \uparrow(\phi_{deconv}(\bm{F}_5))
\label{eq26}
\end{equation}
where the final feature $\bm{F}_6$ can be decoupled as the original motions $\{\bm{M}_{-2},\bm{M}_{-1},\bm{M}_{1},\bm{M}_{2}\}$ as in Eq.\ref{eq1}.

\begin{figure*}[!t]
\centering{
\includegraphics[width=\textwidth]{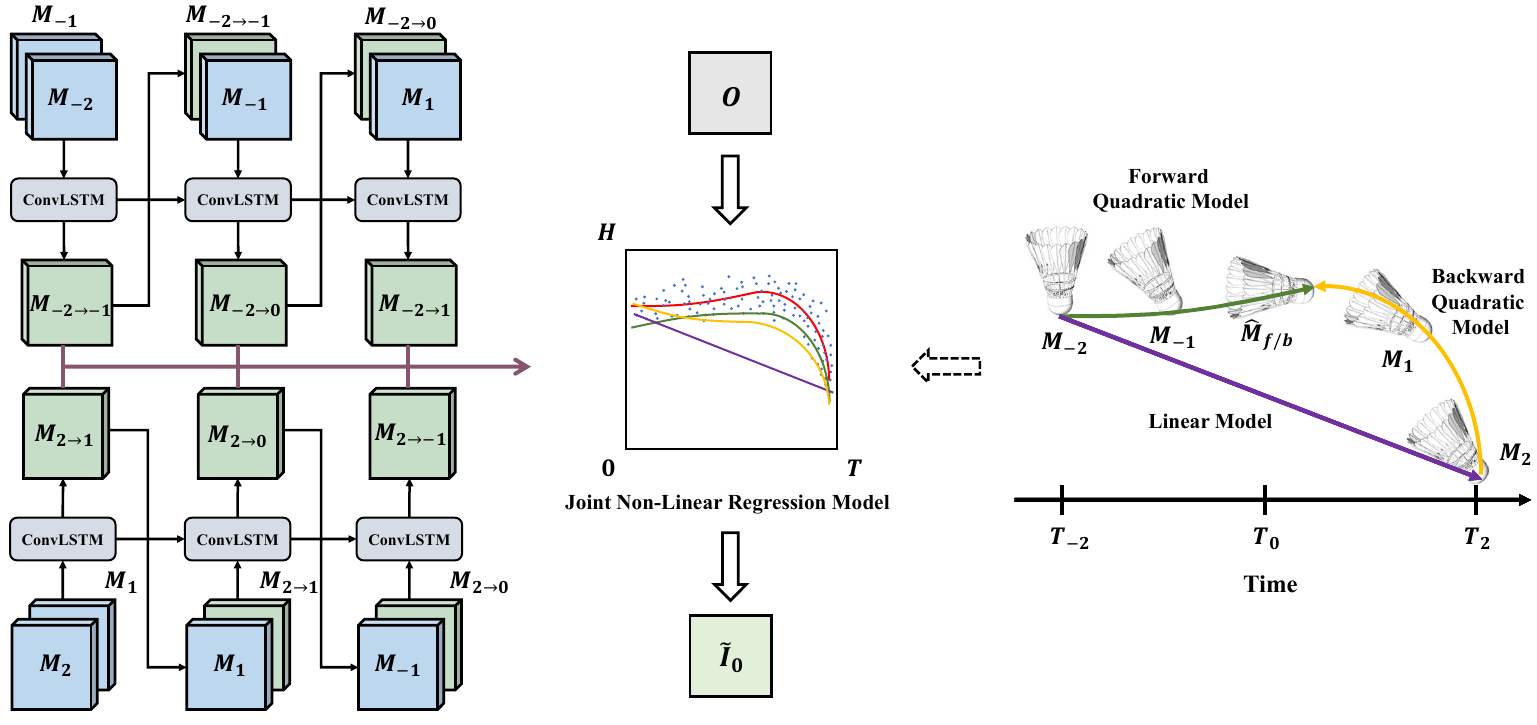}
}
\caption{Illustration of JNMR structure details. Here, we incorporate temporal-aware motion sequences $[\bm{M}_{-2}, \bm{M}_{-1}, \bm{M}_{1}]$ and $[\bm{M}_{2}, \bm{M}_{1}, \bm{M}_{-1}]$ to model different quadratic regressions for the middle motions $[\hat{\bm{M}}_f, \hat{\bm{M}}_b]$. Thus, the entire model is regressed by optimizing each individual sub-non-linear model.}
\label{fig4}
\end{figure*}

\subsection{Joint Non-linear Motion Regression}
\label{secb}
{After obtaining the appropriate motion features $\bm{F}_6$ from the RDFL network, as shown in Fig.\ref{fig3}, joint non-linear motion regression (JNMR) is proposed to implement the aforementioned rationale in Sec.\ref{seca} with four reference motions $\{\bm{M}_{-2},\bm{M}_{-1},\bm{M}_{1},\bm{M}_{2}\}$.} Following Eq.\ref{eq6}, the individual bidirectional regressed motions can be expressed as:
\begin{equation}
\begin{aligned}
\lim_{f\to0^-}\hat{\bm{M}}_{f} &= \frac{(\bm{M}_{1}-\bm{M}_{-1})-2(\bm{M}_{-1}-\bm{M}_{-2})}{3}, \\
\lim_{b\to0^+}\hat{\bm{M}}_{b} &= \frac{(\bm{M}_{-1}-\bm{M}_{1})-2(\bm{M}_{1}-\bm{M}_{2})}{3}
\end{aligned}
\label{eq14}
\end{equation}
where $\hat{\bm{M}}_{f}$ and $\hat{\bm{M}}_{b}$ denote the forward and backward regressed motions in a minimal unilateral neighborhood of the intermediate moment, respectively. {The variation vectors between two frames are temporally combined by the consecutive ConvLSTM elaborated in Fig.\ref{fig4}.} For example, $(\bm{M}_{-2},\bm{M}_{-1})$ and $(\bm{M}_{-1},\bm{M}_{1})$ are input into ConvLSTM to explore relative variation for solving $\hat{\bm{M}}_{f}$ in Eq.\ref{eq14} of the forward temporal dimension. {Later, $\hat{\bm{M}}_{f}$ and $\hat{\bm{M}}_{b}$ are applied in Eq.\ref{eq11} for independent regression to adaptively form a complete regression model.} The regressed expression can be attained as:
\begin{equation}
\begin{split}
    \bm{\hat{\theta}'}&=[\hat{\theta} ~~~ (1-\hat{\theta})], \\
    \bm{\hat{Y}'}&=[ \hat{\bm{M}}_{f} ~~~ \hat{\bm{M}}_{b}]^T
\end{split}
\label{eq15}
\end{equation}
where the regressed coefficient $\hat{\theta}$ is initialized by occlusion $O \in [0,1]$. The visual movement offset $\Delta\hat{I}_0$ can be inferred by Eq.\ref{eq12}, as:
\begin{equation}
\Delta\hat{I}_0 = \hat{\theta} \cdot \varphi(I_{-1}, \hat{\bm{M}}_{f}) + (1-\hat{\theta}) \cdot \varphi(I_{1},\hat{\bm{M}}_{b}) 
\label{eq16}
\end{equation}
As illustrated in Eq.\ref{eq13}, the basic synthesis frame $\hat{I}_0$ can be specifically expressed as:
\begin{equation}
\hat{I}_0=O \cdot (\hat{I}_{-2}+\hat{I}_{-1}) + (1-O) \cdot (\hat{I}_{1} + \hat{I}_{2})
\label{eq17}
\end{equation}
Consequently, the current predicted frame $\tilde{I}_{0}$ can be attained by the combination of $\hat{I}_{0}$ and $\Delta\hat{I}_0$, as:
\begin{equation}
\tilde{I}_{0}=\hat{I}_{0} + \Delta\hat{I}_0
\label{eq18}
\end{equation}

\subsection{Coarse-to-Fine Synthesis Enhancement}
\label{secd}
After obtaining the motions $\{\bm{M}_{-2},\bm{M}_{-1},\bm{M}_{1},\bm{M}_{2}\}$, the interpolation frame $\tilde{I}_0$ is synthesized by JNMR, as illustrated in Sec.\ref{secb}. In general, occlusion affects the reconstructed details and decreases the visual quality of interpolation frames. Therefore, a coarse-to-fine synthesis enhancement (CFSE) module is proposed to further preserve the details of interpolated frames. 

{As described in Sec.\ref{secb} and Fig.\ref{fig3}, the coarse features $\bm{F}_2$ and $\bm{F}_3$ are decoupled into motions, and $\tilde{I}_{0}$ is reconstructed at different scales.} Following GridNet~\cite{fourure2017residual}, the multi-scale reconstructed frames are concatenated to generate a coarse-to-fine interpolated frame $\overline{I}_0$. The final interpolation frame ${I}_0$ is then synthesized with $\tilde{I}_0$ and $\overline{I}_0$, as:
\begin{equation}
{I}_0=\lambda \cdot \tilde{I}_{0} + (1-\lambda) \cdot \overline{I}_{0}
\label{eq21}
\end{equation}
where $\lambda$ denotes the weight coefficient initialized by occlusion. 

\begin{figure*}[!t]
\centering{
\includegraphics[width=\textwidth]{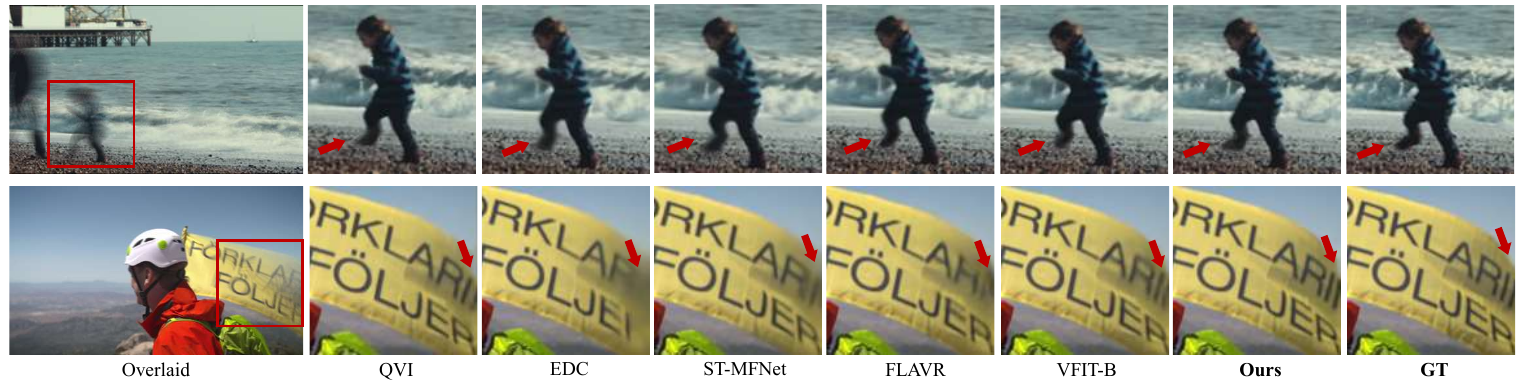}
}
\caption{Qualitative comparisons against the state-of-the-art multi-reference interpolation algorithms with the Vimeo-90K Septuplet dataset. Our method generates higher-quality frames with an exquisite visual appearance and few distortions.}
\label{fig5}
\end{figure*}

\begin{table*}
\setlength\tabcolsep{6pt}
\renewcommand\arraystretch{1.5}
\centering
\footnotesize
\caption{Video frame interpolation: Quantitative comparisons of Vimeo-Septuplet, DAVIS, and GoPro with state-of-the-art methods. The numbers in {\textbf{bold}} and {\underline{underline}} indicate the best and the second best PSNR(dB)/SSIM results with Parameters(\#P.). The run-time (RT) of each model is reported with an input size of 480p frames. The results marked with $\dag$ are cloned from the authors. }\label{tab1}
{
\begin{tabular}{cccccccc}
\hline
{Methods} & {Venue} & {Training Dataset} & {Vimeo-Septuplet} & {DAVIS} & {GoPro}  
&{\#P.(M)} & {RT(s)} \\
\hline   
AdaCoF~\cite{lee2020adacof}  & CVPR'20 & Vimeo-Septuplet   &    34.63         /    0.970  & 27.28  / 0.872   &    29.41      /   0.912  & 21.8 & \textbf{0.04}  \\
CDFI~\cite{ding2021cdfi}    &  CVPR'21  &  Vimeo-Septuplet  &   35.38           /    0.972     &     27.17        /    0.863        &       29.59       /  0.915   & \textbf{5.0} &  0.17      \\
QVI~\cite{xu2019quadratic} & NeurIPS'19 & GoPro    & 35.15           / 0.971   & 27.73         / \underline{0.894}    &      30.55      /     0.933    & 29.2   & \underline{0.16}  \\
$\dag$PRF$_4$-Large~\cite{shen2020video}& TIP'20 & Adobe240   & -    & -   & 31.06             / 0.907           & 11.4    & -       \\
EDC~\cite{danier2022enhancing}& ICIP'22 & Vimeo-Septuplet \& BVI   &       36.14      /  0.974    & 27.63         / 0.878   &     30.93     /     0.936  &  18.5   &  0.35   \\
ST-MFNet~\cite{danier2022spatio} & CVPR'22 &  Vimeo-Septuplet \& BVI    &   36.45          /   {0.976}      &     \underline{28.29}         /  \textbf{0.895} &    30.88           /       0.936     & 21.0   &  0.38       \\
GDConvNet~\cite{Shi2022VideoFrame}& TMM'22 & Vimeo-Septuplet     & 35.58        / 0.958  & 27.03 / 0.821     &     {30.82}     /    {0.913}  & \underline{5.1}  &   0.88   \\
$\dag$NME~\cite{Saikat2022Non} & CVPR'22 & Vimeo-Septuplet   & 34.99          / 0.954       & 27.53            / 0.828 & 29.08             / 0.883    & 20.9   &  -              \\
VFIT-S~\cite{shi2022video} & CVPR'22 & Vimeo-Septuplet    & 36.48           / {0.976}   & 27.92         / 0.885    &    30.55      /  0.939   & {7.5}  &  0.20    \\
VFIT-B~\cite{shi2022video} & CVPR'22 & Vimeo-Septuplet   & {36.96}           /  \underline{0.978}     & 28.09         / 0.888  &    30.60      /  \underline{0.940}  & 29.1   & 0.28     \\
FLAVR~\cite{kalluri2023flavr}& WACV'23 & Vimeo-Septuplet     & 36.30        / 0.975  & 27.44         / 0.874     &     {31.31}     /    \underline{0.940}  & 42.4  &   0.30   \\
$\dag$MA-CSPA~\cite{Zhou2022Exploring}& CVPR'23 &  Vimeo-Septuplet   & 36.50           / 0.962     & -  & -      & 28.9    & -            \\
JNMR(Ours)  & - & Vimeo-Septuplet  & \underline{37.13}   / \underline{0.978}    & {28.25}             / 0.887    &      \underline{32.46}        /      \textbf{0.951}     & 5.7   & 0.38 \\
JNMR(Ours)  & - & Vimeo-Septuplet \& BVI  & \textbf{37.19}   / \textbf{0.979}    & \textbf{28.32}           / {0.889}    &      \textbf{32.47}        /      \textbf{0.951}     & 5.7   & 0.38\\
\hline
\end{tabular}
}
\end{table*}

\subsection{Objective Function}
For the end-to-end training, we utilize the objective, perceptual and deformation loss to measure the difference between the synthesized frame ${I}_0$ and its ground truth $I_{gt}$. Specifically, the $\ell_1$ loss with the Charbonnier penalty~\cite{liu2017video} is introduced as:
\begin{equation}
\mathcal{L}_{Charbon}=\rho({I}_0-I_{gt})
\label{eq22}
\end{equation}
where $\rho(x)=(||x||_2^2+\epsilon^2)^{\frac{1}{2}}$ and $\epsilon=0.001$.  

The perceptual loss $\mathcal{L}_{vgg}$ is expressed as: 
\begin{equation}
\mathcal{L}_{vgg}=||\Phi({I}_0)-\Phi(I_{gt})||_2
\label{eq23}
\end{equation}
where $\Phi ( )$ is a feature extraction from conv4\_3 of the pre-trained VGG16~\cite{simonyan2014very}.

The deformation loss $\mathcal{L}_d$ is implemented to measure the abnormal deformation distance of each directional motion vector at the kernel-level as follows: 

\begin{equation}
\mathcal{L}_d=\sum_{i,j}||(\bm{\alpha},\bm{\beta})_{i,j+1}-(\bm{\alpha},\bm{\beta})_{i,j}||_1+||(\bm{\alpha},\bm{\beta})_{i+1,j}-(\bm{\alpha},\bm{\beta})_{i,j}||_1
\label{eq24}
\end{equation}

\noindent where $(\alpha,\beta)_{i,j}$ represents the regressed kernel-level motion vectors in $\hat{\bm{M}}_n$ and bidirectional motions in $\hat{\bm{M}}_f$ and $\hat{\bm{M}}_b$.

Three loss functions are combined to optimize the whole network parameters for end-to-end training as follows:
\begin{equation}
\mathcal{L}=\mathcal{L}_{Charbon}+\lambda_{vgg}\mathcal{L}_{vgg}+\lambda_d\mathcal{L}_d
\label{eq25}
\end{equation}
where $\lambda_{vgg}$ and $\lambda_d$ denote the weight coefficients of perceptual loss and deformation loss, respectively.

\section{Experiments}

\subsection{Implementation Details}
\subsubsection{Configuration} {All experiments are implemented on two NVIDIA GeForce RTX 3090 GPUs with Intel(R) Xeon(R) Gold 6226R CPUs. We conduct 120 training epochs with the Vimeo-90K Septuplet~\cite{xue2019video} dataset and 30 fine-tuning epochs on BVI-DVC~\cite{ma2021bvi} dataset with a mini-batch size of 8. AdaMax~\cite{kingma2015adam} is the optimizer with $\beta_1=0.9$ and $\beta_2=0.999$. The learning rate is set as $1e^{-3}$ to $1.5e^{-5}$ by half decay every 20 epochs.} $\lambda_{vgg}$ and $\lambda_d$ are set to 0.005 and 0.01, respectively.
\subsubsection{Training Datasets} We train our model using the Vimeo-90K Septuplet~\cite{xue2019video} dataset, which includes 64,612 and 7,824 seven-frame sequences with a resolution of $256\times448$. {To further improve the performance for large motions, we use the BVI-DVC~\cite{ma2021bvi} dataset, which includes 17,600 quintuplets with a resolution of $256\times256$, to fine-tune for better evaluation.} The middle frame of each septuplet and quintuplet is the interpolation target, and its adjacent four consecutive frames are used as the input frames in Fig.\ref{fig3}. {We also apply random horizontal, vertical flipping, and temporal order reversal to further enhance the training dataset.}
\subsubsection{Evaluation Datasets} The experimental model is evaluated not only on the validation set of the Vimeo-90K Septuplet but also on other commonly used benchmark datasets, such as DAVIS~\cite{perazzi2016benchmark} and GoPro~\cite{nah2017deep} as previously demonstrated in QVI~\cite{xu2019quadratic}. {Using the same sampling principle, we report PSNR and SSIM~\cite{wang2004image} with 2,849 quintuples generated from DAVIS and 3,012 quintuples with a resized resolution of $480\times854$ from GoPro.}

\begin{figure*}[!t]
\centering{
\includegraphics[width=\textwidth]{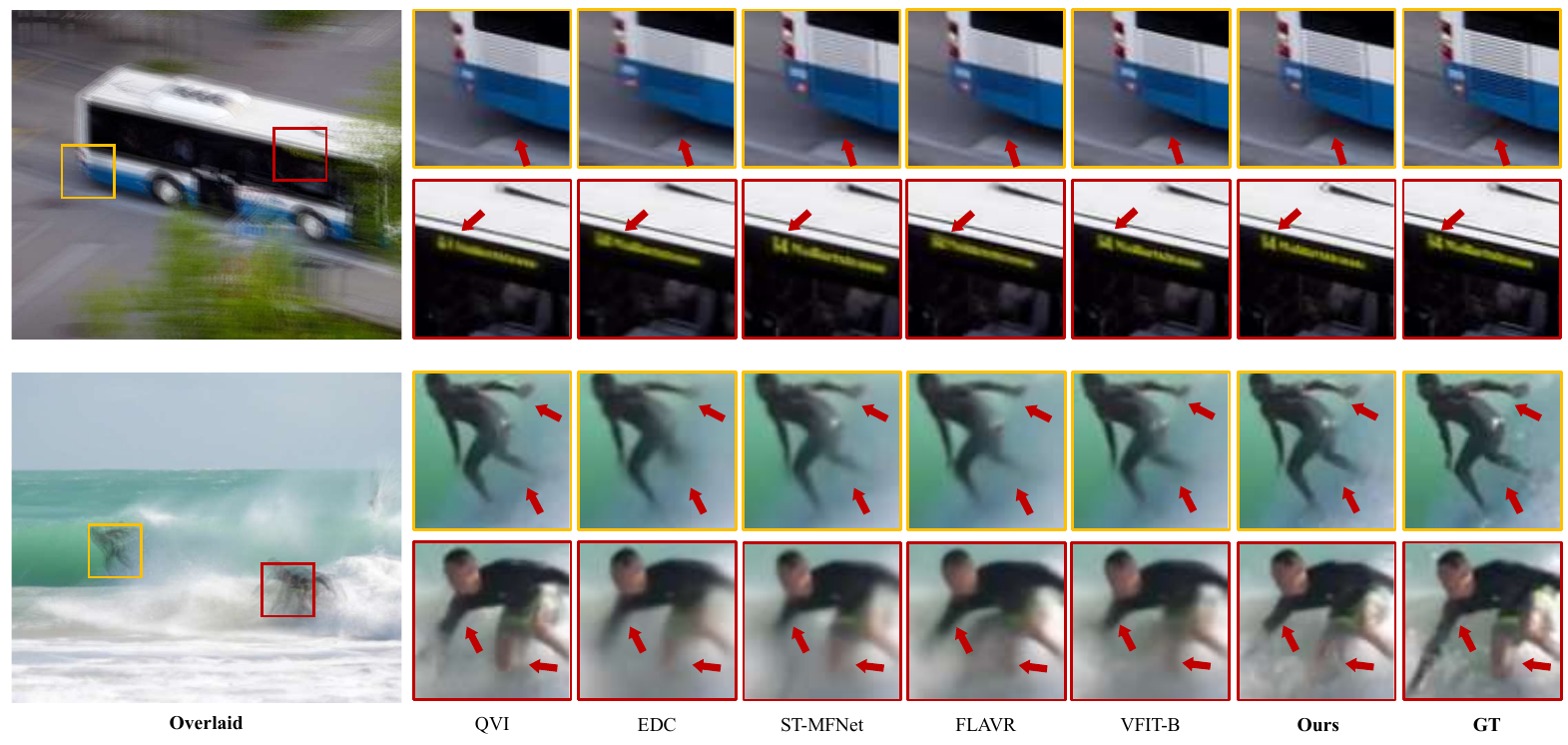}
}
\caption{Qualitative evaluation of the high-speed motion of the DAVIS dataset. JNMR not only outperforms other models in terms of texture reconstruction and artifact reduction but also generates a reasonable middle frame with structural consistency when dealing with large motions.}
\label{fig6}
\end{figure*}

\subsection{Evaluation against State-of-the-art Methods}

{To verify the effectiveness of JNMR, we make a fair comparison with state-of-the-art methods using four reference frames, such as QVI~\cite{xu2019quadratic}, PRF$_4$~\cite{shen2020video}, EDC~\cite{danier2022enhancing}, ST-MFNet~\cite{danier2022spatio}, GDConvNet~\cite{Shi2022VideoFrame},  FLAVR~\cite{kalluri2023flavr}, MA-CSPA~\cite{Zhou2022Exploring}, NME~\cite{Saikat2022Non} and VFIT~\cite{shi2022video}. With regard to QVI, EDC, ST-MFNet, GDConvNet, FLAVR and VFIT, the pre-trained models are directly used with the same experimental setups for evaluation.} {In addition, we compare AdaCoF~\cite{lee2020adacof} and CDFI~\cite{ding2021cdfi} using two reference frames with the Vimeo-90K Septuplet.} Regarding other studies presenting methods without publicly available code\footnote{The results are marked by $\dag$ in TABLE~\ref{tab1}.}, we conduct a comparison with the results kindly provided by the study authors.

\subsubsection{Quantitative Evaluation}
As shown in TABLE~\ref{tab1}, our proposed JNMR has great superiority with the Vimeo-90K Septuplet, DAVIS, and GoPro benchmarks with exceptional performance in terms of model parameters and running time. Notably, JNMR achieves an interpolation performance beyond 37 dB with the Vimeo-90K Septuplet dataset. JNMR further improves the interpolation performance without complicated feature learning and synthesis by exploiting the temporal-aware acceleration information through joint regression. {Moreover, after the fine-tuning process with the BVI-DVC dataset followed by the ST-MFNet, the experimental results in TABLE~\ref{tab1} show that JNMR has improved on different test datasets.} {With only 5.7M parameters, JNMR promotes significant improvements over state-of-the-art methods, such as 0.23 dB with the Vimeo-90K Septuplet and 1.16 dB with GoPro.} In summary, our JNMR method achieves new state-of-the-art performance with strong generalization for different benchmarks with competitive parameters.

\begin{figure}
\centering{
\includegraphics[width=\linewidth]{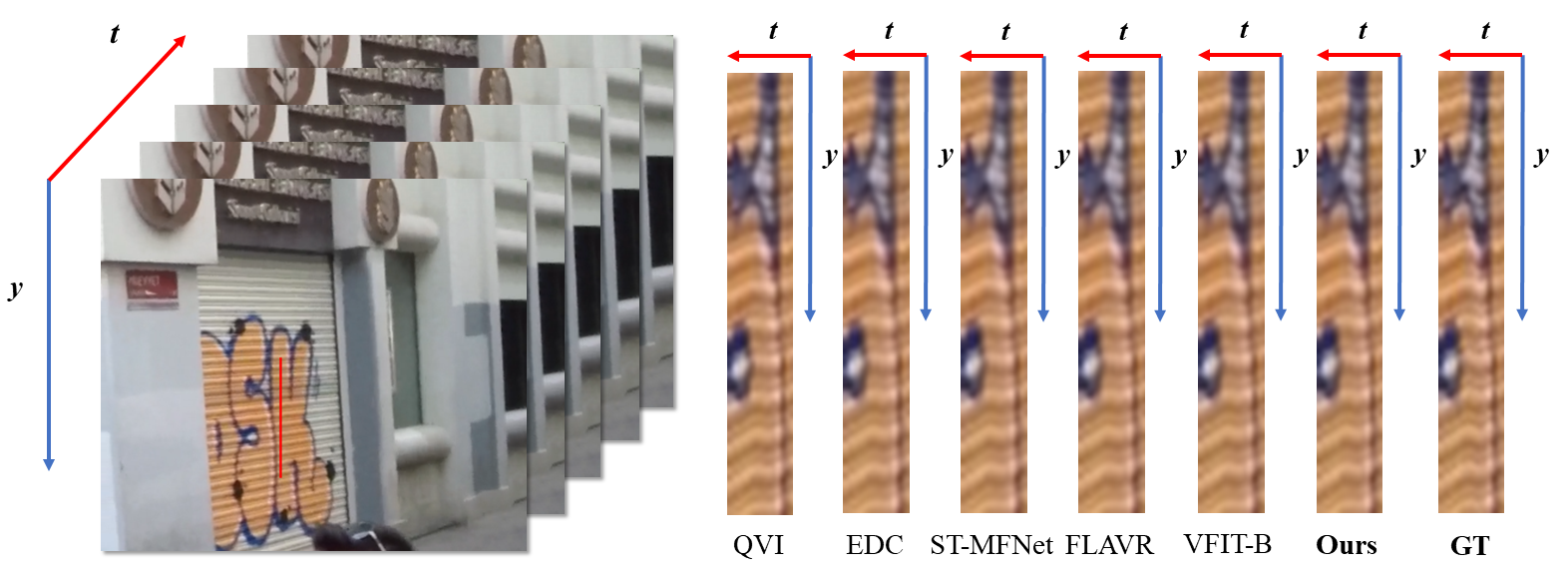}
}
\caption{The temporal profile on the GroPro dataset. The temporal profile is generated by stacking a single-pixel line (red light) among video sequences. The resource sequence contains fifteen consecutive frames, where the nine middle frames are generated by different models.}
\label{fig7}
\end{figure}

\subsubsection{Qualitative Evaluation}
We provide a qualitative comparison between our method and the latest state-of-the-art methods with the Vimeo-90K Septuplet and DAVIS datasets in Fig.\ref{fig5} and Fig.\ref{fig6}. As shown in Fig.\ref{fig5}, JNMR generates visually pleasing interpolation results with clear structures on Vimeo-90K Septuplet. The two samples both contain complicated motion with low resolution. In particular, the second sample involves structural details such as characters. {It is noted that other methods fail to restore the original appearance of the child's foot and the rightmost text.} In contrast, our model achieves a sharp boundary and realistic texture without excessive smoothing due to the appropriate feature learning and motion regression. Furthermore, we demonstrate the temporal-aware performance and the reconstruction effect on high-speed movement in Fig.\ref{fig6}. Although EDC~\cite{danier2022enhancing}, ST-MFNet~\cite{danier2022spatio}, FLAVR~\cite{kalluri2023flavr} and VFIT-B~\cite{shi2022video} generate visually correct structures, there is some serious blurring due to the direct fusion of overlaid input. QVI~\cite{xu2019quadratic} relies on the quadratic modeling of motions, but their interpolation frames usually contain notable artifacts because of inaccurate synthesis. In contrast, our method successfully handles complicated acceleration movement modeling and produces plausible structures with abundant details. In particular, the relative position between the cement marking line and the bus in the first sample indicates the temporal consistency in consecutive frames. {JNMR can realize a close relation to the ground truth (GT) with clear edges and demonstrate effective capabilities in handling temporal-aware high-speed motions. In addition, it is noted that our method can achieve clear text reconstruction and edge preservation of high-speed moving objects.} 

\begin{figure*}[!t]
\centering{
\includegraphics[width=\textwidth]{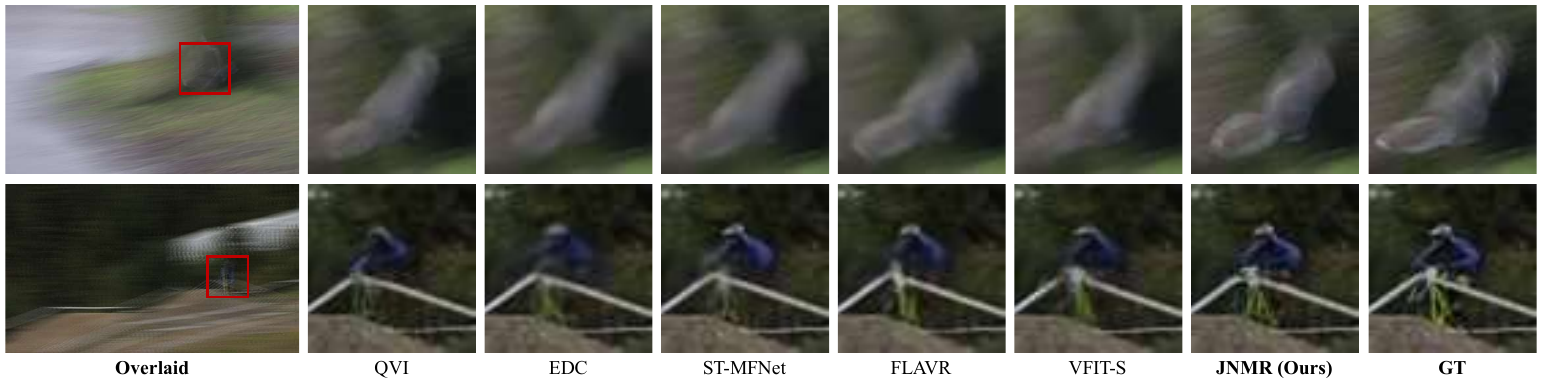}
}
\caption{Qualitative comparisons against the state-of-the-art multi-reference interpolation algorithms with the Vimeo-90K Septuplet dataset. The middle frames generated by JNMR have clear high-frequency structures and few artifacts compared with other models when dealing with large and complicated motions.}
\label{fig11}
\end{figure*}

\begin{table}
\setlength\tabcolsep{6pt}
\renewcommand\arraystretch{1.3}
\centering
\footnotesize
\caption{Quantitative comparisons on the slow, medium and fast sub-collection of the Vimeo-90K Septuplet with state-of-the-art methods. The numbers in {\textbf{bold}} and {\underline{underline}} indicate the best and the second best PSNR(dB)/SSIM results, respectively.}\label{tab7}
{
\begin{tabular}{cccc}
\hline
\multirow{2}{*}{Methods} & \multicolumn{3}{c}{Vimeo-90K Septuplet} \\
\cline{2-4} 
& {Slow} & {Medium} & {Fast}   \\
\hline
AdaCoF~\cite{lee2020adacof} & 37.54 / 0.982 & 34.55 / 0.970 & 31.11 / 0.946 \\
CDFI~\cite{ding2021cdfi} & 38.27 / 0.984 & 35.23 / 0.973 & 32.19 / 0.951 \\
QVI~\cite{xu2019quadratic}   &  37.99 / 0.982 &     35.04 / 0.971   &     32.08 / 0.949      \\
EDC~\cite{danier2022enhancing}  &  38.71 / {0.987}  &  36.08  / 0.975 &    32.99 / 0.952      \\
ST-MFNet~\cite{danier2022spatio}   &  38.70 / {0.987} &   36.35  / 0.976  &    33.88  /  {0.957}     \\
GDConvNet~\cite{Shi2022VideoFrame}  & 38.01  / 0.978 &    35.58 / 0.961  &    32.36  / 0.923      \\
FLAVR~\cite{kalluri2023flavr}  & 38.72  / \underline{0.988}  &   36.25  /  \underline{0.977} &   33.34   /  0.953     \\
VFIT-S~\cite{shi2022video}&  \underline{38.73} / {0.987} & \underline{36.40} / \underline{0.977}  &  \underline{33.88} /  \underline{0.958}     \\
JNMR(Ours)   &  \textbf{39.47} / \textbf{0.989}   &   \textbf{37.12}  / \textbf{0.978}    &    \textbf{34.45}  /  \textbf{0.959}       \\
\hline
\end{tabular}
}
\end{table}

\subsubsection{Temporal Consistency}
{We apply the temporal consistency~\cite{song2021multi} to measure the movement of pixel lines and further evaluate the structural continuity in interpolated videos. Methods that exhibit similar pixel tendencies as the ground truth are deemed to preserve the original temporal variations effectively.} We present the visual results of the comparison methods on the GoPro dataset in Fig.\ref{fig7}. {The ground truth contains crooked horizontal lines that indicate camera platform movement. Our method reflects these dynamic scenes with a fine high-frequency context. Other methods produce overly smooth results and fail to capture the turning point of pixel direction under long-term dynamics.} From the above evaluation, it can be seen that JNMR is an effective method for restoring subtle temporal variations using motion regression.

\subsubsection{Evaluation on Large and Complicated Motions}
To verify the superiority of JNMR on large and complicated motions, we evaluate our method with Vimeo-90K Septuplet sub-collections according to the methods of previous work~\cite{Haris2019Recurrent}. The test sequences are stratified into slow, medium and fast sub-collections by estimated motion velocities. Compared to the other state-of-the-art methods with similar complexity, as shown in TABLE~\ref{tab7}, our method performs better performance on different motion velocities. {Specifically, it can be seen that motion regression among multiple reference frames can effectively improve VFI performance compared with our baseline methods.} {As shown in Fig.\ref{fig11}, the qualitative example on the fast sub-collection demonstrates that JNMR does not produce more motion artifacts when dealing with complicated and large motions, especially in the case of irregular camera movement in the first example.}

\subsection{Ablation Study}
{In this section, {we present the results of the comprehensive ablation studies to evaluate the contribution of the JNMR strategy and other auxiliary sub-components with the Vimeo-90K Septuplet dataset.} The quantitative evaluation results of individual components with the baseline model are shown in TABLE~\ref{tab2}.}

\begin{table}
\setlength\tabcolsep{4pt}
\renewcommand\arraystretch{1.3}
\centering
\footnotesize
\caption{Ablation results of individual sub-component. }\label{tab2}
\begin{tabular}{cccccc}
\hline
Models &   \#P.(M)    & PSNR(dB)       & SSIM         \\
\hline   
Baseline    & 5.3   &     36.82        &   0.975             \\
Baseline w/ RDFL     &  \textbf{4.0}  &      36.82       &     0.975           \\
Baseline w/ JNMR     &  6.9  &    37.08(+0.26)        &     0.978(+0.003)     \\
Baseline w/ CFSE     &  5.4  &     36.98(+0.16)       &     0.976(+0.001)  \\
JNMR(Full)     & 5.7 & \textbf{37.19}(+0.37)         & \textbf{0.979}(+0.004)                  \\
\hline
\end{tabular}
\end{table}

\subsubsection{Joint Non-linear Motion Regression}
\begin{table}
\setlength\tabcolsep{4pt}
\renewcommand\arraystretch{1.3}
\centering
\footnotesize
\caption{Ablation results of different regression models.}\label{tab3}
\begin{tabular}{c||c||cccc}
\hline
Models &  Illustrations & \#P.(M)    & PSNR(dB)        & SSIM         \\
\hline   
 {Model 1}    &   {Linear }   &        \textbf{4.1}      &  36.95            &   0.976    \\
 {Model 2}      &   {Quadratic }  &  5.7   &  37.04  & 0.977           \\
 {Model 3}      &   {Linear combination of quadratic}  &   5.7   &  37.04   &         0.977    \\
 {Model 4}      & {Unidirectional}   &     5.7        &     37.10    & 0.978         \\
 {Model 5}    &   {Second-order unidirectional }  &     5.7        &     37.05    &     0.977       \\

 {JNMR}     &  {Joint bidirectional} &  5.7 &  \textbf{37.19}           &  \textbf{0.979}                  \\
\hline
\end{tabular}
\end{table}

\begin{figure*}[!t]
\centering{
\includegraphics[width=\textwidth]{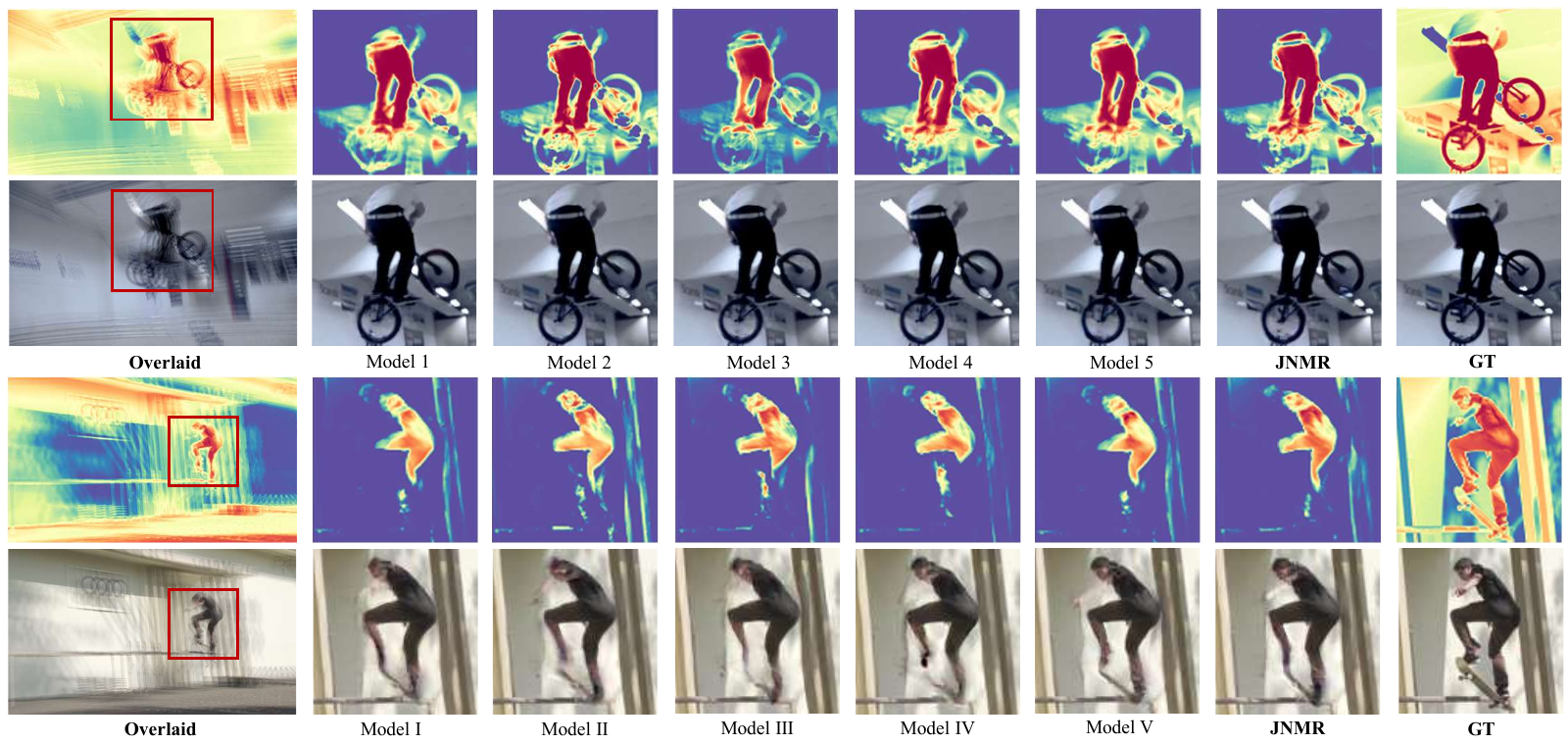}
}
\caption{Qualitative comparisons with the different regression models and reconstruction strategies with the Vimeo-90K Septuplet dataset. It is noted that JNMR generates reasonable motion with an exquisite visual appearance.}
\label{fig9}
\end{figure*}

The proposed JNMR is designed to model variable acceleration curve motion as a multi-stage quadratic movement. {To verify the effectiveness of this strategy, we conduct different multi-variable regressions as shown in TABLE~\ref{tab3}. We first compare the performance of the linear (Model 1) and quadratic (Model 2) models. Then, the validity of the temporal-aware combination using ConvLSTM illustrated in Sec.\ref{secb} is verified. We use a linear combination of quadratic models (Model 3) to derive $\hat{M}_f$ and $\hat{M}_b$ without ConvLSTM. Model 3 can also be seen as the natural cubic spline. Both Model 3 and natural cubic spline obtain a quadratic curve locally and sequentially process the video sequence. We also compare two different regression strategies, unidirectional regression and second-order unidirectional regression. In particular, unidirectional regression (Model 4) only uses forward/backward motions in Eq.\ref{eq15}, such as $\bm{\hat{Y}'} = [\hat{\bm{M}}_{f}]^T$ or $\bm{\hat{Y}'} = [\hat{\bm{M}}_{b}]^T$. The second-order unidirectional regression (Model 5) makes a key ablation in terms of regression direction through serial repeating of the above pipeline with both directions.}

\begin{table}
\setlength\tabcolsep{4pt}
\renewcommand\arraystretch{1.3}
\centering
\footnotesize
\caption{Ablation results on different numbers of hierarchical structures with multi-stage compensation. }\label{tab4}
\begin{tabular}{c||cc||ccc}
\hline
\multirow{2}{*}{Models}  
&\multirow{2}{*}{\begin{tabular}[c]{@{}c@{}}Structures\\ Number\end{tabular}} 
&\multirow{2}{*}{\begin{tabular}[c]{@{}c@{}}Multi-stage\\ Compensation\end{tabular}} 
& \multirow{2}{*}{\#P.(M)}
& \multirow{2}{*}{PSNR(dB)}
& \multirow{2}{*}{SSIM}          \\
& & & & & \\
\hline   
{Model \uppercase\expandafter{\romannumeral1}}    &   5   &        \ding{56}      &        7.0     &  37.11 & 0.978  \\
{Model \uppercase\expandafter{\romannumeral2}}    &   3   &        \ding{56}      &        \textbf{5.6}     &  37.05 & 0.978    \\
{JNMR}      &   3  &      \ding{52}       &     5.7 & \textbf{37.19} & \textbf{0.979}           \\
\hline
\end{tabular}
\end{table}

\begin{table}
\setlength\tabcolsep{4pt}
\renewcommand\arraystretch{1.3}
\centering
\footnotesize
\caption{The ablation results on different source features for coarse-to-fine synthesis enhancement module.}\label{tab5}
\begin{tabular}{c||cc||cccc}
\hline
Models &  Source Features &  GridNet & \#P.(M)    & PSNR(dB)       & SSIM         \\
\hline   
{Model \uppercase\expandafter{\romannumeral3}}    &   - & \ding{56}  &        \textbf{5.6}      &        37.03     &  0.977    \\
{Model \uppercase\expandafter{\romannumeral4}}    &   {$\bm{F}_1,  \bm{F}_2$}   &  \ding{52}  &    5.7     &        37.06     &  0.977    \\
{Model \uppercase\expandafter{\romannumeral5}}    &   {$\bm{F}_2,  \bm{F}_3$}   &  \ding{56}  &    \textbf{5.6}     &        37.06     &  0.977    \\
{JNMR}      &   {$\bm{F}_2,  \bm{F}_3$}  &  \ding{52} &    5.7         &  \textbf{37.19}        & \textbf{0.979}      \\
\hline
\end{tabular}
\end{table}

The evaluation results demonstrate the robustness of our ratiocination as shown in TABLE~\ref{tab3} and Fig.\ref{fig9}. Our regression strategy successfully restores the correct occlusion and consistent patterns in handling complicated motions of the sample. Moreover, the visualization of $\tilde{I}_0$ verifies that JNMR can interpolate motion with clear edge details and few artifacts. Notably, our reliable kinematic model is robust for different dynamic visual scenes.

\subsubsection{Feature Learning and Frame Synthesis}
As described in Sec.\ref{sec3}, {an appropriate network architecture is explored to retain semantic information favorable for motion regression.} The coarse-to-fine synthesis enhancement module is implemented to preserve finer details, with the different resolution motions integrated into the final frame. {TABLE~\ref{tab4} shows the performance of the network with different numbers of hierarchical spatial structures illustrated in Eq.\ref{eq20}. Model I, which has five hierarchical spatial structures, does not outperform the JNMR. It indicates that more complex feature extraction structures are not very helpful for frame reconstruction and motion regression but also bring about an increase in parameters. In addition, the multi-stage compensation strategy with a few parameters improves the performance compared to Model II which has only up-sampling operations.} To verify the efficiency of the coarse-to-fine synthesis enhancement module, we conduct an ablation study on source features of different resolutions. As described in TABLE~\ref{tab5}, {the multi-stage extraction influences the visual details laterally. Besides, GridNet has proven effective in multi-scale feature fusion through Model \uppercase\expandafter{\romannumeral5}. In addition to the advantages in quantitative evaluation, our method can also restore comprehensive structures, in contrast to the other ablation methods shown in Fig.\ref{fig9}.}

\begin{figure}[!t]
\centering
\setlength{\abovecaptionskip}{0.cm}
{
\includegraphics[width=\linewidth]{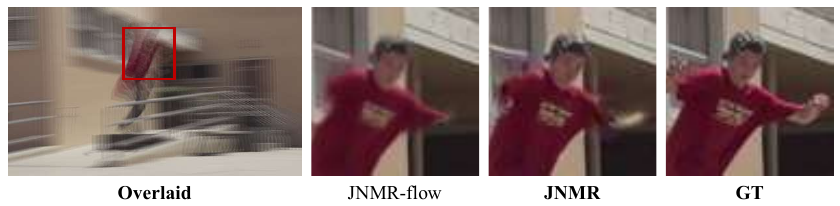}
}
\caption{Qualitative comparisons with different motion estimation methods with the Vimeo-90K Septuplet dataset. Flow-based motions are able to reduce artifacts but extract less complicated motion information.}
\label{fig12}
\end{figure}

\begin{table}
\setlength\tabcolsep{4pt}
\renewcommand\arraystretch{1.3}
\centering
\footnotesize
\caption{Ablation results on different motion estimation methods.}\label{tab8}
\begin{tabular}{c||c||cccc}
\hline
Models &  Motion Estimation Methods  & \#P.(M)    & PSNR(dB)       & SSIM         \\
\hline   
{JNMR-flow}    &   Optical Flow   &        9.5       &        36.96     &  0.976    \\
{JNMR}      &   Deformable Convolution  &    \textbf{5.7}         &  \textbf{37.19}        & \textbf{0.979}      \\
\hline
\end{tabular}
\end{table}

\begin{figure*}
\centering{
\includegraphics[width=\linewidth]{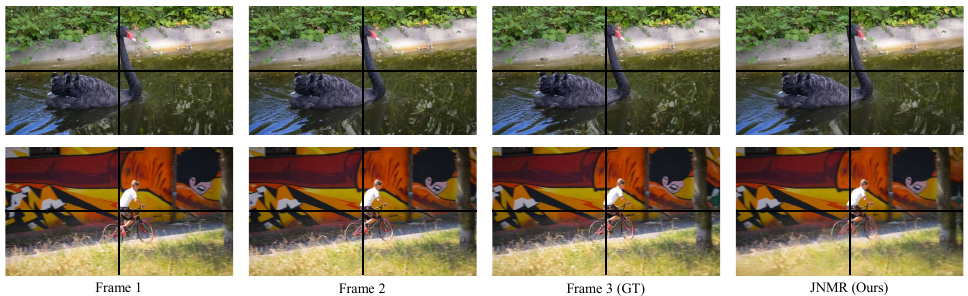}
}
\caption{{The qualitative evaluation of extrapolation on the DAVIS dataset. The moving targets in the frame sequence are indicated by the calibration of the black cross.}}
\label{fig10}
\end{figure*}

\begin{table}[!t]
\setlength\tabcolsep{4pt}
\renewcommand\arraystretch{1.3}
\centering
\footnotesize
\caption{Ablation results on the different numbers of reference frames with the GoPro dataset.}\label{tab9}
\begin{tabular}{c||c||cccc}
\hline
Models &  Reference Number  & \#P.(M)    & PSNR(dB)       & SSIM         \\
\hline   
JNMR  & 4 & {5.7}       & {32.47}   &   {0.951}  \\
JNMR-6f & 6  & {6.0}     & {32.52}   &   {0.952}  \\
\hline
\end{tabular}
\end{table}

\subsubsection{Flow-based Motion Estimation}
To further verify the universality of our method, the pre-trained LiteFlowNet~\cite{hui2018liteflownet} is utilized to generate optical flow motions instead of deformable convolution denoted as JNMR-flow. JNMR-flow can also achieve good VFI performance, but uses more model parameters, as illustrated in TABLE~\ref{tab8}. However, the whole network is mainly designed toward the feature-level, which leads to the lower interpolation performance of JNMR-flow compared to the kernel-based JNMR. In addition, Fig.\ref{fig12} indicates that the pixel-level optical flow of JNMR-flow is still inferior to the feature-level offset map of JNMR in catching large motions.

\subsubsection{The Number of Reference Frames}
{To validate the influence of the reference frame number, we input 6 reference frames to interpolate the intermediate frame, denoted as JNMR-6f. Particularly, the first and last frames of 4-frame input and 6-frame input from GoPro dataset are constant in different models. Due to the increase in the number of reference frames, the intermediate motion estimation is more accurate. Therefore, as illustrated in TABLE~\ref{tab9}, the results of JNMR-6f are slightly higher than those of the 4 reference frames (JNMR). The two models remain consistent overall and have similar performance, demonstrating the generalization of JNMR on long sequences.}

\begin{table}[!t]
\setlength\tabcolsep{4pt}
\renewcommand\arraystretch{1.3}
\centering
\footnotesize
\caption{Video frame extrapolation: Quantitative comparisons with the Vimeo-Triplet and Adobe240 with state-of-the-art methods. The numbers in {\textbf{bold}} and {\underline{underline}} indicate the best and the second best PSNR(dB) and SSIM results with Parameters(\#P.). The results of other methods are cloned from \cite{liu2020convtransformer} and \cite{Zhou2022Exploring}. }\label{tab6}
{
\begin{tabular}{ccccccc}
\hline
\multirow{2}{*}{Methods} 
&\multirow{2}{*}{\begin{tabular}[c]{@{}c@{}}\#P.\\ (M)\end{tabular}} & 
\multicolumn{2}{c}{Vimeo-Triplet} & \multicolumn{2}{c}{Adobe240}\\
\cline{3-6}
&           & PSNR        & SSIM        & PSNR      & SSIM                         \\
\hline   
Convtransformer~\cite{liu2020convtransformer}     & -  & 30.52           &   0.941  &       \underline{30.42}       & \underline{0.946}          \\
DVF~\cite{liu2017video}   &  \textbf{3.8}  &  27.08   &      0.907      &    28.74   &      0.925               \\
MCNet~\cite{villegas2017decomposing}   &   -  &        28.62       &  0.873   &        28.21     &  0.880                      \\
Sepconv~\cite{niklaus2017video}  &  21.7   &     30.42          &  0.917   &     -       &  -                      \\
FLAVR~\cite{kalluri2023flavr}  &  42.1  &       31.14        &  0.927   &     -         &  -                      \\
MA-CSPA~\cite{Zhou2022Exploring}  &  22.4   &       \textbf{32.05}        &  \underline{0.940}   &    -         &  -                       \\
JNMR (Ours)  & \underline{4.5}     & \underline{31.55}           & \textbf{0.947}    & \textbf{31.33}             &   \textbf{0.952}                   \\
\hline
\end{tabular}
}
\end{table}

\subsection{Extension for Extrapolation}
To further explore the extension of the motion regression, we follow the work DVF~\cite{liu2017video} to conduct JNMR in the video frame extrapolation to generate future frames with several reference frames. {In detail, we predict the next frame utilizing two consecutive frames and make a quantitative evaluation with the Vimeo-Triplet and Adobe240~\cite{su2017deep} datasets.} As shown in TABLE~\ref{tab6}, JNMR achieves improvements on most evaluation indicators with fewer parameters. {The qualitative evaluation results on the DAVIS dataset are shown in Fig.\ref{fig10}. The predicted objects in JNMR have the same relative position compared to ground truth, which demonstrates the effectiveness of JNMR in maintaining the temporal consistency of moving objects.} 

\section{Conclusion}
{In this paper, we conduct an analysis on the importance of long-term dynamics in the task of video frame interpolation.}  To overcome the challenge of large and complicated motion synthesis, a joint non-linear motion regression (JNMR) strategy is designed to introduce multi-variate non-linear regression for interpolation. {Our method formulates the kinematic trajectory as joint multi-stage quadratic models and achieves accurate and consistent motion prediction.} Furthermore, regression-driven feature learning and coarse-to-fine synthesis enhancement modules are explored to maintain global structures and complement details for regression. {The experimental results demonstrate the superior performance and robustness of JNMR compared to other state-of-the-art methods.}

\bibliographystyle{IEEEtran}
\bibliography{egbib}
\balance
\begin{IEEEbiography}
[{\includegraphics[width=1in,height=1.25in,clip,keepaspectratio]
{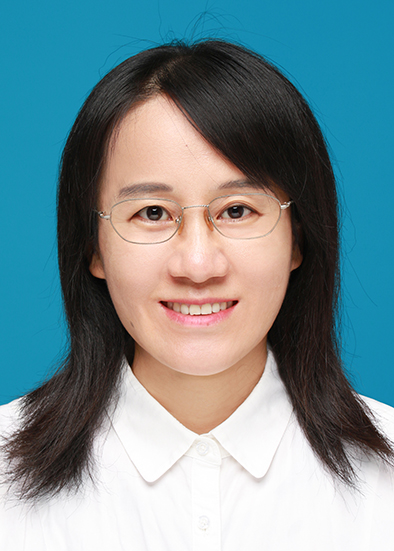}}]{Meiqin Liu} received the M.E. degree and Ph.D. degree from Beijing Jiaotong University (BJTU), China, in 2007 and 2018, respectively. From 2014 to 2015, she was a Visiting Scholar at Simon Fraser University (SFU), Canada. She is currently an Associate Professor at the Institute of Information and Science, BJTU. Her research interests include image/video compression and video processing.
\end{IEEEbiography}
\vspace{-0.3 cm}
\begin{IEEEbiography}
[{\includegraphics[width=1in,height=1.25in,clip,keepaspectratio]{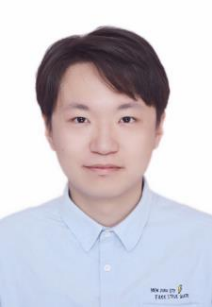}}] 
{Chenming Xu} received the B.S. degree from China University of Geosciences (CUG), Wuhan, China, in 2021. He is currently pursuing the M.E. degree at the Institute of Information Science, Beijing Jiaotong University (BJTU), China. 
His research interests include video restoration and video compression.
\end{IEEEbiography}
\vspace{-0.3 cm}
\begin{IEEEbiography}
[{\includegraphics[width=1in,height=1.25in,clip,keepaspectratio]{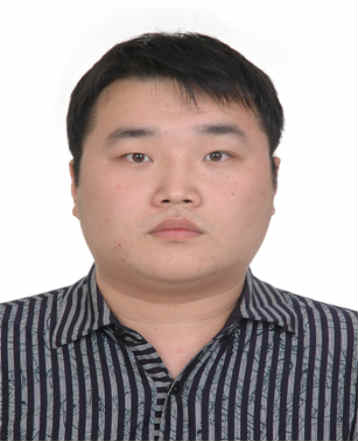}}] 
{Chao Yao} received the M.E. degree and Ph.D. degree from Beijing Jiaotong University (BJTU) in 2010 and 2016. From 2014 to 2015, he was a Visiting Ph.D. student with LTS4 Group, Institute of the Swiss Federal Institute of Technology (EPFL), Lausanne, Switzerland. He is currently an Associate Professor with University of Science and Technology Beijing (USTB). His research interests include image/video compression, computer vision and human-computer interaction. 
\end{IEEEbiography}
\begin{IEEEbiography}
[{\includegraphics[width=1in,height=1.25in,clip,keepaspectratio]{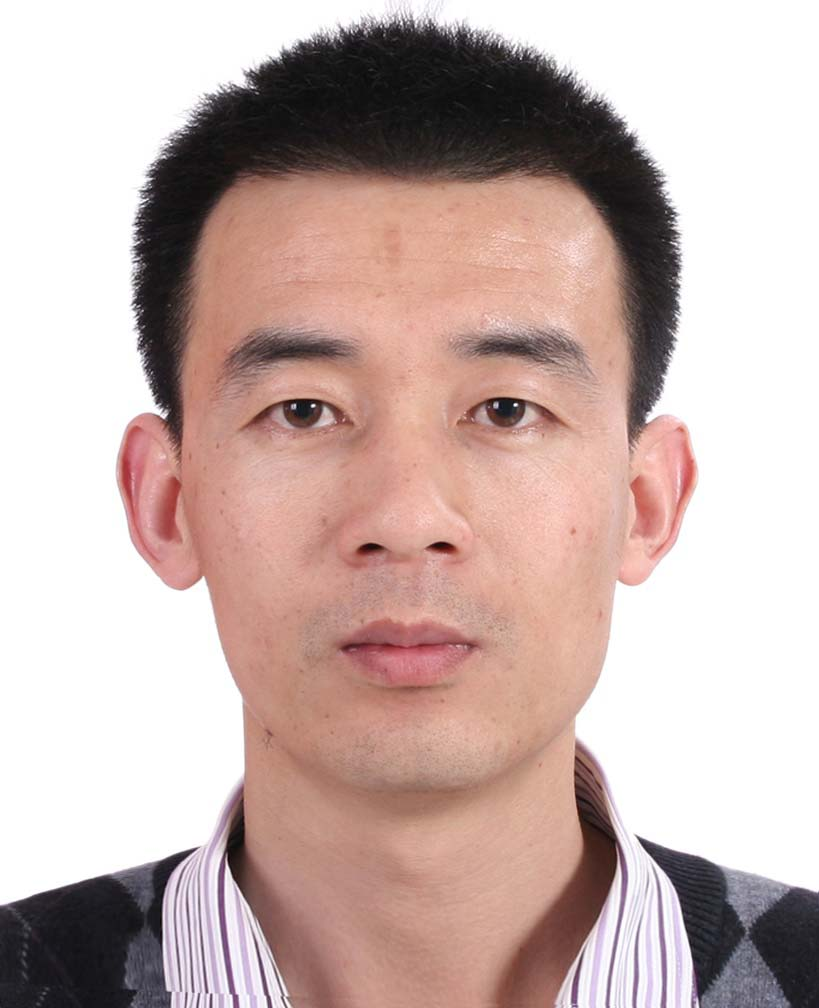}}] 
{Chunyu Lin} (Member, IEEE) received the Ph.D. degree from Beijing Jiaotong University (BJTU), Beijing, China, in 2011. From 2009 to 2010, he was a Visiting Researcher with the ICT Group, Delft University of Technology, The Netherlands. From 2011 to 2012, he was a Postdoctoral Researcher with the Multimedia Laboratory, Gent University, Belgium. He is currently a Professor with BJTU. His research interests include image/video compression and robust transmission, 3D vision, virtual reality video processing, and ADAS.
\end{IEEEbiography}
\begin{IEEEbiography}
[{\includegraphics[width=1in,height=1.25in,clip,keepaspectratio]{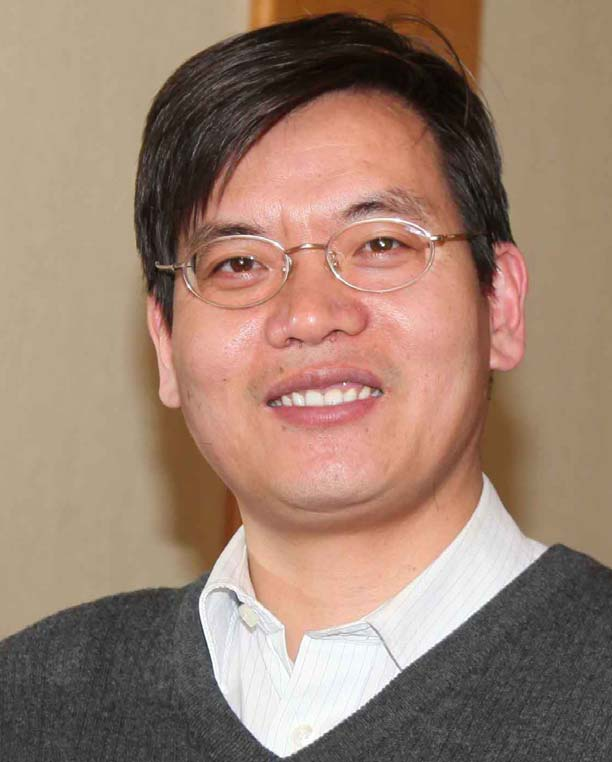}}] 
{Yao Zhao} (Fellow, IEEE) received the B.S. degree from the Radio Engineering Department, Fuzhou University, Fuzhou, China, in 1989, the M.E. degree from the Radio Engineering Department, Southeast University, Nanjing, China, in 1992, and the Ph.D. degree from the Institute of Information Science, Beijing Jiaotong University (BJTU), Beijing, China, in 1996. He is currently the Director of the Institute of Information Science, Beijing Jiaotong University. His current research interests include image/video coding and video analysis and understanding. He was named a Distinguished Young Scholar by the National Science Foundation of China in 2010 and was elected as a Chang Jiang Scholar of Ministry of Education of China in 2013.
\end{IEEEbiography}
\end{document}